\documentclass{article}

\PassOptionsToPackage{numbers, compress}{natbib}

\usepackage[final]{neurips_2022}

\usepackage[utf8]{inputenc} \usepackage[T1]{fontenc}    \usepackage{hyperref}       \usepackage{url}            \usepackage{booktabs}       \usepackage{amsfonts}       \usepackage{nicefrac}       \usepackage{microtype}      \usepackage{xcolor}         \usepackage{graphicx}
\usepackage{xspace}
\usepackage[font=small]{caption}
\usepackage{wrapfig}
\usepackage{amsmath}
\usepackage{algorithm}
\usepackage{algorithmicx}
\usepackage{algpseudocode}
\usepackage{multirow}
\usepackage{array}
\usepackage{enumitem}
\usepackage[numbers]{natbib}
\usepackage{csquotes}
\usepackage[group-separator={,},group-minimum-digits={3}]{siunitx}

\newcommand{\ours}[0]{\textsc{DreamGrader}\xspace}
\newcommand{\dream}[0]{\textsc{Dream}\xspace}
\newcommand{\program}[0]{\mu}

\newcommand{\icon}[1]{\includegraphics[scale=0.36]{images/icons/icon_#1.pdf}}
\newcommand{\rl}{RL$^2$\xspace}

\title{Giving Feedback on Interactive Student Programs with Meta-Exploration}

\author{Evan Zheran Liu\thanks{Co-first authors. Correspondence to \texttt{evanliu@cs.stanford.edu}.} , Moritz Stephan$^*$\!, Allen Nie, Chris Piech, Emma Brunskill, Chelsea Finn \\
   Stanford University
}

\begin{document}

\maketitle

\vspace{-2mm}
\begin{abstract}
    Developing interactive software, such as websites or games, is a particularly engaging way to learn computer science.
    However, teaching and giving feedback on such software is time-consuming --- standard approaches require instructors to manually grade student-implemented interactive programs.
    As a result, online platforms that serve millions, like Code.org, are unable to provide any feedback on assignments for implementing interactive programs, which critically hinders students' ability to learn.
One approach toward automatic grading is to learn an agent that interacts with a student's program and explores states indicative of errors via reinforcement learning.
However, existing work on this approach only provides binary feedback of whether a program is correct or not, while students require finer-grained feedback on the specific errors in their programs to understand their mistakes.
    In this work, we show that exploring to discover errors can be cast as a meta-exploration problem.
    This enables us to construct a principled objective for discovering errors and an algorithm for optimizing this objective, which provides fine-grained feedback.
We evaluate our approach on a set of over 700K real anonymized student programs from a Code.org interactive assignment.
    Our approach provides feedback with 94.3\% accuracy, improving over existing approaches by 17.7\% and coming within 1.5\% of human-level accuracy. Project web page: \url{https://ezliu.github.io/dreamgrader}.
\end{abstract}

\section{Introduction}
\vspace{-4mm}

\begin{wrapfigure}[19]{r}{.6\textwidth}
    \vspace{-6.5mm}
    \includegraphics[width=0.6\textwidth]{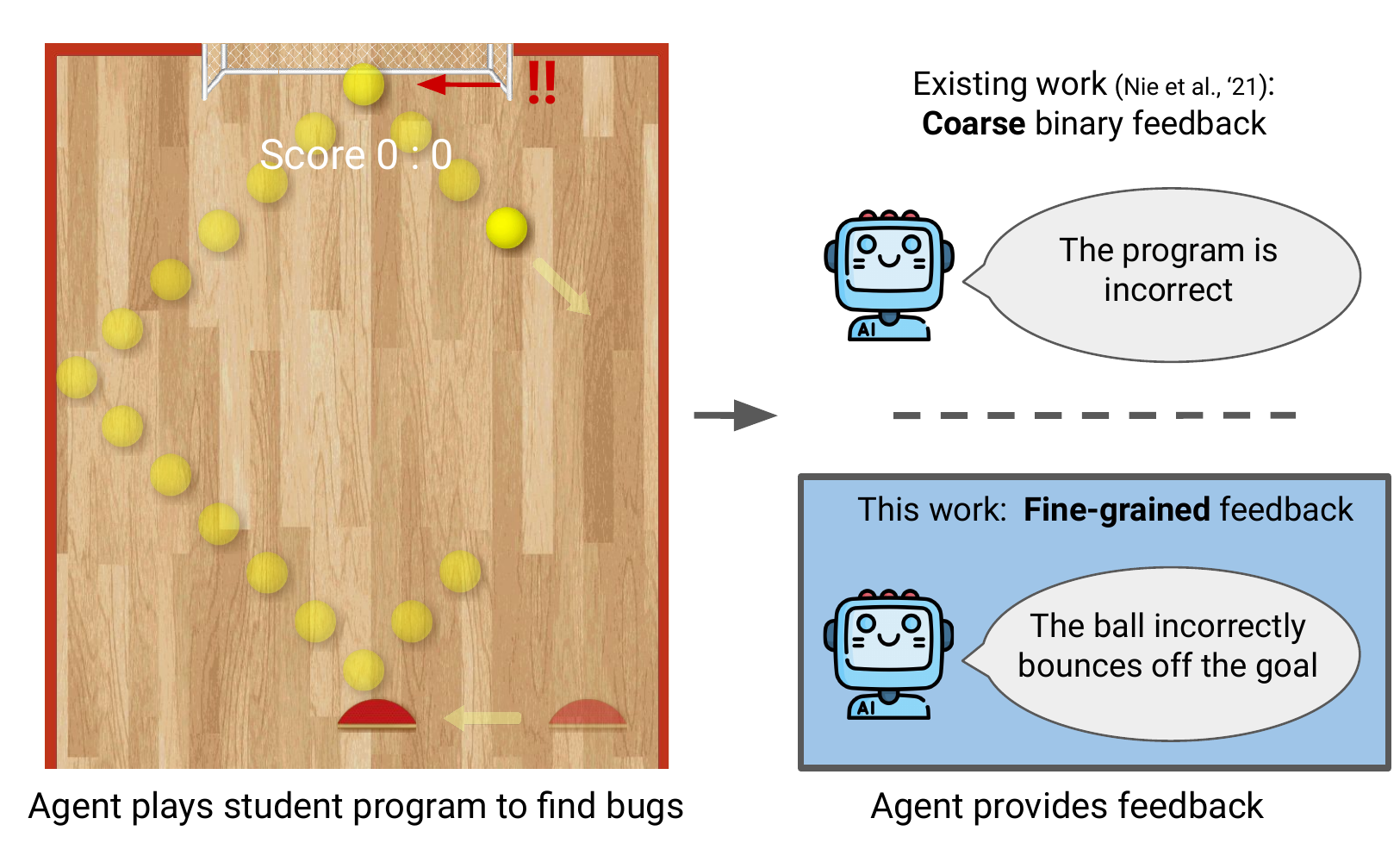}
    \vspace{-4.5mm}
    \caption{
        A learned Play-to-Grade agent for the Bounce programming assignment. The agent tests what happens when the ball is hit into the goal, and finds that the ball incorrectly bounces out instead of scoring a point. Whereas prior work provides coarse feedback of whether the program is correct or not, our goal is to provide fine-grained feedback on the specific mistakes a student has made.
    }\label{fig:overview}
\end{wrapfigure}

\looseness=-1
Feedback plays a critical role in high-quality education, but can require significant time and expertise to provide~\citep{code2022org}.
We focus on one area where providing feedback is particularly burdensome: modern computer science education, where students are often tasked with developing interactive programs, such as websites or games (e.g., see Figure~\ref{fig:overview}).
While developing such programs is highly engaging~\citep{pfaffman2003manipulating} and has become ubiquitous in contemporary classrooms~\citep{froyd2012five}, these programs can include stochastic or creative elements, so they cannot be graded with traditional unit tests and must instead be manually graded.
However, such manual grading is increasingly infeasible with the growing demand for computer science education and rise of massive online learning platforms. For example, one popular platform, Code.org has enrolled
over 70M students~\citep{code2022org}. As manually grading a submission can take up to 6 minutes, a single assignment creates decades of grading labor. Consequently, platforms like Code.org cannot yet provide feedback about whether an interactive assignment submission is correct or not, let alone more fine-grained feedback.

\looseness=-1
To alleviate this enormous grading burden, \citet{nie2021play} introduce the Play-to-Grade paradigm for automatically providing feedback by training a reinforcement learning agent to grade a program the same way humans do: by interacting or playing with the program.
The idea is for the agent to visit states that reveal errors in the program, and then aggregate this information as feedback.
Such an agent is trained on a set of training programs labeled with feedback (e.g., provided by an instructor), and the goal is to generalize to new student programs. Figure~\ref{fig:overview} shows an example learned agent that tests what happens when the ball is hit into the goal, exposing an error where the ball bounces off the goal instead of entering and scoring a point. The state-of-the-art approach in this paradigm provides accurate coarse feedback of whether the program is completely correct or not~\cite{nie2021play}. However, to understand their mistakes, students usually require more specific feedback about what errors are in their programs.

\looseness=-1
Learning an agent to explore and discover the errors in a program to provide such fine-grained feedback is challenging:
Most errors cannot be discovered with simple random exploration, and instead require targeted exploration, such as deliberately hitting the ball into the goal.
In addition, the agent must be able to adapt its exploration to different programs, which each behave differently and may present unexpected obstacles, such as multiple balls.
Our key technical insight is that learning to discover errors connects with the meta-exploration problem in meta-reinforcement learning (meta-RL).
This insight enables us to leverage techniques from the meta-exploration literature to construct and optimize a principled objective for producing fine-grained feedback.
Specifically, we follow the Play-to-Grade paradigm and assume access to 3,556 training programs, labeled with the errors in the program.
Then, we formulate the problem as maximizing the mutual information between states visited by our agent and the label.
Finally, we use techniques from the \dream meta-RL algorithm~\citep{liu2021dream} to decompose this objective into shaped rewards that enable learning sophisticated exploration.

\looseness=-1
Overall, the main contribution of this work is to connect the Play-to-Grade paradigm with the meta-RL literature, and consequently, to provide an effective method for providing fine-grained feedback for interactive student programs, which we call \ours.
Additionally, we release our code to propose automatic feedback as a new meta-RL testbed that fulfills an unmet need in the community for a benchmark that is simultaneously readily accessible and directly impactful.
We evaluate our system on 711,274 anonymized student submissions of the Bounce assignment from Code.org \citep{nie2021play}.
Trained on 3,556 programs, \ours achieves an accuracy of 94.3\%, which improves over existing approaches by 17.7\% and comes within 1.5\% of human-level grading accuracy.
In addition, our approach can significantly reduce instructor grading burden: while manually grading all student submissions would require
4 years of work, our system can grade the same set of programs $180\times$ faster on a single GPU and can be parallelized for even faster grading over multiple GPUs.

\section{Related Works}\label{sec:related_work}

\looseness=-1
\textbf{Educational feedback.} We consider the problem of automatically providing feedback, which plays an important role in student learning and motivation~\citep{orourke2014hint}.
Though we specifically focus on feedback, other works also leverage machine learning for other aspects of education, including tracking what students know~\citep{villano1992probabilistic, d2008more, piech2015deep}, predicting student retention~\citep{delen2011predicting, balakrishnan2013predicting, sass2018structural, aulck2016predicting}, and building intelligent tutoring systems~\citep{anderson1990cognitive, nwana1990intelligent}.
Work on automatically providing feedback for computer science assignments focuses on two main approaches: analyzing either (i) the code or (ii) the behavior of a program.
Methods that analyze code provide feedback by passing the code through a neural network~\citep{piech2015learning,bhatia2016automated,malik2019generative,wu2019zero},
or by constructing syntax trees~\citep{singh2013automated,wang2017data}, e.g., to predict useful next implementation steps~\citep{rivers2017data,paassen2017continuous}. Analyzing code works well for shorter programs (e.g., under 50 lines of code) and has even been deployed in online courses~\citep{wu2021prototransformer}. However, this approach struggles to scale to lengthier or more complex programs. Hence, we instead opt for the second approach of analyzing program behavior, which conveniently does not depend on program length, though it requires the program to be executable.

\looseness=-1
Arguably, the simplest method of analyzing program behavior is unit testing. Unit testing can provide automatic feedback to some extent when the desired output of each input is known, but this is typically not the case with interactive programs, such as websites or games. Instead, work on automated testing can provide feedback by generating corner-case inputs that reveal errors via input fuzzing~\citep{godefroid2008automated}, symbolic generation~\citep{king1976symbolic}, or reinforcement learning exploration objectives~\citep{zheng2019wuji,gordillo2021improving}. However, this line of work assumes that errors are easy to detect when revealed, while detecting a revealed error itself can be challenging~\citep{nie2021play}.

Consequently, \citet{nie2021play} propose the Play-to-Grade paradigm to both learn an agent to discover states that reveal errors, and a model to detect and output the revealed errors. Our work builds upon the Play-to-Grade paradigm, but differs from \citet{nie2021play} in the provided feedback. While \citet{nie2021play} only provide coarse binary feedback of whether a program is correct, we introduce a new principled objective to provide fine-grained feedback of what specific errors are present to help students understand their mistakes.

\textbf{Meta-reinforcement learning.} To provide fine-grained feedback, we connect the problem of discovering errors with the meta-exploration problem in meta-RL. There is a rich literature of approaches that learn to explore via meta-RL~\citep{gupta2018meta,stadie2018importance,rakelly2019efficient,rothfuss2018promp,zhou2019environment,zintgraf2019varibad,humplik2019meta,gurumurthy2019mame,kamienny2020learning,liu2021dream}. We specifically leverage ideas from the \dream algorithm~\citep{liu2021dream} to construct a shaped reward function for learning exploration.
Our work has two key differences from prior meta-RL research. First, we introduce a novel factorization of the \dream objective that better generalizes to new programs. Second, and more importantly, we focus on the problem of providing feedback on interactive programs. This differs from a large body of meta-RL work that focuses on interesting, yet synthetic problems, such as 2D and 3D maze navigation~\citep{duan2016rl,mishra2017simple,zintgraf2019varibad,liu2021dream}, simulated control problems~\citep{finn2017modelagnostic,rakelly2019efficient,yu2019metaworld}, and alchemy~\citep{wang2021alchemy}. While meta-RL has been applied to realistic settings in robotics~\citep{nagabandi2018learning,arndt2020meta,schoettler2020meta}, such application requires costly equipment.
In contrast, this work provides a new meta-RL problem that is both realistic and readily accessible.

\section{The Fine-Grained Feedback Problem}\label{sec:problem_setting}

We consider the problem of automatically providing feedback on programs.
During training, we assume access to a set of programs labeled with the errors made in the program (i.e., ground-truth instructor feedback).
During testing, the grading system is presented with a new student program and must output feedback of what errors are in the program.
To produce this feedback, the grading system is allowed to interact with the program.

More formally, we consider a distribution over programs $p(\program)$, where each program $\program$ defines a Markov decision process (MDP) $\program = \langle \mathcal{S}, \mathcal{A}, \mathcal{T}, \mathcal{R}\rangle$ with states $\mathcal{S}$, actions $\mathcal{A}$, dynamics $\mathcal{T}$, and rewards $\mathcal{R}$.
We assume that the instructor creates a \emph{rubric}: an ordered list of $K$ potential errors that can occur in a program.
Each program $\program$ is associated with a ground-truth label $y \in \{0, 1\}^K$ of which errors are made in the program.
The $k^\text{th}$ index $y_k$ denotes that the $k^\text{th}$ error of the rubric is present in the program $\program$.

During training, the grading system is given a set of $N$ labeled training programs $\{(\program^n, y^n)\}_{n = 1}^N$.
The goal is to learn a \emph{feedback function} $f$ that takes a program $\program$ and predicts the label $\hat{y} = f(\program)$ to maximize the \emph{expected grading accuracy} $\mathcal{J}_\text{grade}(f)$ over test programs $\program$ with \emph{unobserved} labels $y$:
\begin{equation}\label{eqn:grading_accuracy}
    \mathcal{J}_\text{grade}(f) = \mathbb{E}_{\program \sim p(\program)}\left[\frac{1}{K}\sum_{k = 1}^K \mathbb{I}[f(\program)_k = y_k] \right],
\end{equation}
where $\mathbb{I}$ is an indicator variable.
Effectively, $\mathcal{J}_\text{grade}$ measures the per-rubric item accuracy of predicting the ground-truth label $y$.
To predict the label $y$, the feedback function may optionally  interact with the MDP $\program$
defined by the program for any small number of episodes.

\looseness=-1
\textbf{Bounce programming assignment.}
Though the methods we propose in this work generally apply to any interactive programs with instructor-created rubrics and we include experiments on another interactive assignment in Appendix~\ref{app:breakout}, we primarily focus on the Bounce programming assignment from Code.org, a real online assignment that has been completed nearly a million times.
As providing feedback for interactive assignments is challenging,
the assignment currently provides no feedback whatsoever on Code.org, and instead relies on the student to discover their own mistakes by playing their program.
This assignment is illustrated in Figure~\ref{fig:overview}.
Each student program defines an MDP,
where we use the state representation from \citet{nie2021play}: each state consists of the $(x, y)$-coordinates of the paddle and balls, as well as the $(x, y)$-velocities of the balls.
There are three actions: moving the paddle left or right, or keeping the paddle in the current position.
In the dynamics of a correct program, the ball bounces off the paddle and wall.
When the ball hits the goal or floor, it disappears and launches a new ball, which increments the player score and opponent score respectively.
However, the student code may define other erroneous dynamics, such as the ball passing through the paddle or bouncing off the goal.
The reward is $+1$ when the player score increments and $-1$ when the opponent score increments.
An episode terminates after $100$ steps or if either the player or opponent score exceeds $30$.

Our experiments use a dataset of 711,274 real anonymized student submissions to this assignment, released by \citet{nie2021play}.
We use $0.5\%$ of these programs for training, corresponding to $N = \num{3556}$ and uniformly sample from the remaining programs for testing.

\begin{wraptable}[8]{r}{0.6\textwidth}
    \center\small
    \vspace{-6mm}
    \caption{Possible event and consequence types of program errors.
    }
    \vspace{-2mm}
    \resizebox{0.6\textwidth}{!}{\begin{tabular}{cp{1mm}c}
        \toprule
        \textbf{Event} &  & \textbf{Consequence} \\
        \cmidrule(lr){1-1} \cmidrule(lr){3-3}
        Ball hits paddle & {\Large \multirow{6}{*}{\!\!$\times$}} & \multirow{6}{*}{\shortstack{Ball bounces / does not bounce \\ Increments / does not increment player score \\ Increments / does not increment opponent score \\ Launches / does not launch a new ball \\ Moves the paddle}} \\
        Ball hits wall & & \\
        Ball hits goal & &  \\
        Ball hits floor & & \\
        Paddle moves & &  \\
        Program starts & &  \\
        \bottomrule
    \end{tabular}
    }
    \label{tab:error_types}
    \vspace{-3.5mm}
\end{wraptable}

Possible errors in a student program take the form of ``when \emph{event} occurs, there is an incorrect \emph{consequence},'' where the list of all events and consequences is listed in Table~\ref{tab:error_types}.
For example, Figure~\ref{fig:overview} illustrates the error where the event is the ball hitting the goal, and the consequence is that the ball incorrectly bounces off the goal, rather than entering the goal.
For simplicity, we primarily consider a representative rubric of $K = 8$ errors, spanning all event and consequence types, listed in Appendix~\ref{sec:dataset_details}, though we include some experiments on all error types in Appendix~\ref{sec:additional_results}.

\looseness=-1
\textbf{Prior approach for program feedback.}
Accurately determining which errors are in a given program is challenging, because it requires targeted exploration that adapts to variability in the programs.
Often, the presence of some errors makes it difficult to find other errors, such as multiple balls making it difficult to determine which events change the score.
Prior work by \citet{nie2021play} sidesteps this challenge by only providing coarse feedback of whether a program is correct or not, by determining if a student program differs from a reference solution program.
Such coarse feedback is much easier to provide, as it often involves only finding the most obvious error, which can frequently be found with relatively untargeted exploration.
In the next section, we present a new approach that instead targets exploration toward uncovering specific misconceptions to effectively provide fine-grained feedback.
We discuss how our approach differs from~\citet{nie2021play} in greater detail in Appendix~\ref{sec:detailed_prior_work}.

\section{Automatically Providing Fine-Grained Feedback with \ours}\label{sec:approach}

In this section, we detail our approach, \ours, for automatically providing fine-grained feedback to help students understand their mistakes.
From a high level, \ours learns two components that together form the feedback function $f$:

\begin{enumerate}[label=(\roman*), leftmargin=*]
    \item An \emph{exploration policy} $\pi$ that acts on a program $\program$ to produce a trajectory $\tau~=~(s_0, a_0, r_0, \ldots)$.
\item A \emph{feedback classifier} $g(y \mid \tau)$ that defines a distribution over labels $y$ given a trajectory $\tau$.
\end{enumerate}

The idea is to explore states that either indicate or rule out errors with the exploration policy, and then summarize the discovered errors with the feedback classifier.
To provide feedback on a new program $\program$, we first roll out the exploration policy $\pi$ on the program to obtain a trajectory $\tau$, and then obtain the predicted label $\arg\max_y g(y \mid \tau)$ by applying the feedback classifier.
Under this parametrization of the feedback function $f$, we can rewrite the expected grading accuracy objective in Equation~\ref{eqn:grading_accuracy} as:
\begin{align}\label{eqn:reparam_grading_accuracy}
    \mathcal{J}_\ours(\pi, g) = \mathbb{E}_{\program \sim p(\program), \tau \sim \pi(\program)}\left[ \frac{1}{K} \sum_{k = 1}^K \mathbb{I}[\arg\max_{\hat{y}} g(\hat{y} \mid \tau)_k = y_k] \right],
\end{align}
where $\pi(\program)$ denotes the distribution over trajectories from rolling out the policy $\pi$ on the program $\program$.

After this rewriting of the objective, our approach is conceptually straightforward: we learn both the exploration policy and classifier to maximize our rewritten objective.
We can easily learn the feedback classifier $g$ by maximizing the probability of the correct label given a trajectory generated by the exploration policy with standard supervised learning (i.e., cross-entropy loss),
but learning the exploration policy $\pi$ is more challenging.
Note that we could directly optimize our objective in Equation~\ref{eqn:reparam_grading_accuracy} by treating the inside of the expectation as a reward received at the end of the episode and use this to learn the exploration policy $\pi$ with reinforcement learning.
However, this reward signal makes credit assignment difficult for learning the exploration policy,
since it is given at the end of the episode, rather than at the states where the exploration policy discovers errors.
Indeed, we empirically find that learning from this reward signal struggles to adequately explore (Section~\ref{sec:experiments}).

Hence, our goal is instead to construct a reward signal that helps assign credit for the exploration policy and provides rewards at the states that indicate or rule out errors in the program.
To do this, we first propose an alternative objective that is sufficient to maximize the Play-to-Grade objective, but can be decomposed into per-timestep rewards that correctly assign credit (Section~\ref{sec:dream_grader}).
Intuitively, these rewards leverage the feedback classifier to provide high reward when an action leads to a state that increases the classifier's certainty about whether an error is present.
Then, we detail practical design choices for implementing this approach with neural networks (Section~\ref{sec:practical_impl}) and conclude by drawing a connection between learning to find errors and the meta-exploration problem, which motivates our choice of alternative objective and its subsequent decomposition (Section~\ref{sec:metarl}).

\subsection{Assigning Credit to Learn Exploration}\label{sec:dream_grader}

\begin{figure}
    \vspace{-7mm}
    \centering
    \includegraphics[width=\linewidth]{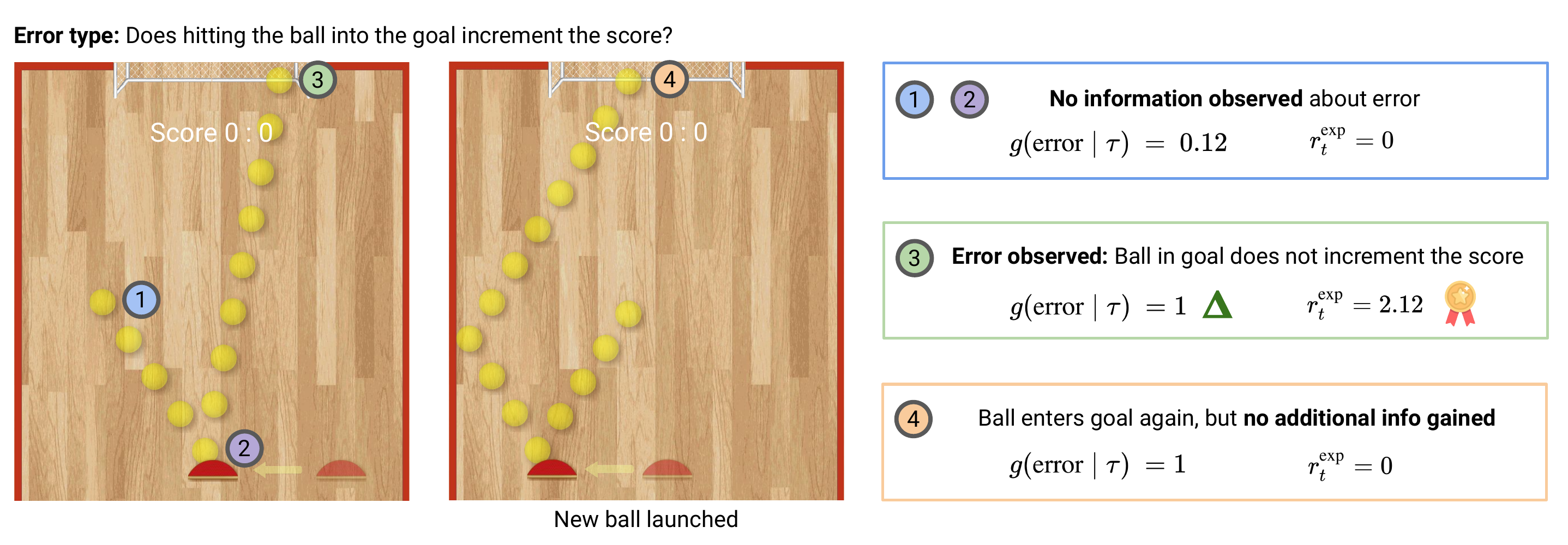}
    \vspace{-4mm}
    \caption{
        \ours provides credit assignment for learning exploration by leveraging the feedback classifier $g(y \mid \tau)$.
        Here, we consider the error ``when the ball enters the goal, the player score does not increment.''
        At \icon{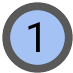} and \icon{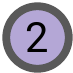}, no information is observed that either rules out or indicates the error.
        Hence, no exploration reward $r^\text{exp}_t$ is provided, and the classifier assigns $0.12$ probability that the error is present, reflecting the prior that 12\% of the training programs have this error.
        At \icon{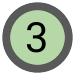}, the ball enters the goal, but does not score a point and a new ball is launched.
        This indicates that the error is present, so the classifier updates, which creates high exploration reward and credit assignment for learning exploration.
        At \icon{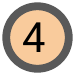}, the ball enters the goal again.
        However, no additional information is gained, so the classifier does not change, and no reward is given.
        Overall, this enables learning effective exploration that purposely hits the ball into the goal once.
    }\label{fig:approach}
    \vspace{-4mm}
\end{figure}

We now obtain a reward signal to help assign credit for learning exploration.
From a high level, we first propose to maximize a mutual information objective that is sufficient to maximize our objective $\mathcal{J}_\ours(\pi, g)$ in Equation~\ref{eqn:reparam_grading_accuracy}.
We then rewrite our mutual information objective in terms of per-timestep exploration rewards related to the information gain of the feedback classifier on the true label $y$ when it observes the transition $(s_t, a_t, r_t)$.
This helps assign credit for learning the exploration policy, as the transitions that either indicate or rule out errors in the program are exactly those that have high information gain.
Figure~\ref{fig:approach} illustrates an example of the derived exploration rewards.

\textbf{Objective.}
Intuitively, we want our exploration policy to visit states that either indicate or rule out errors.
We can formalize this intuition by maximizing the mutual information $I(\tau; y)$ between the trajectories $\tau \sim \pi$ visited by the policy and the feedback label of the program $y$, which causes the policy to visit states that make the feedback label highly predictable.
Importantly, maximizing this objective is sufficient to maximize the expected grading accuracy $\mathcal{J}_{\ours}(\pi, g)$ in Equation~\ref{eqn:reparam_grading_accuracy}:
Let $p(y \mid \tau)$ be the true posterior over labels given trajectories $\tau \sim \pi$ sampled from the policy.
Maximizing the mutual information $I(\tau; y)$ produces trajectories that maximize the probability of the label under the true posterior $p(y \mid \tau)$.
Then, if the feedback classifier is learned to match the true posterior $g(y \mid \tau) = p(y \mid \tau)$ while $I(\tau; y)$ is maximized, the expected grading accuracy $\mathcal{J}_\ours(\pi, g)$ is also maximized.

\textbf{Optimization.}
We can efficiently maximize our objective $I(\tau; y)$
by maximizing a variational lower bound~\citep{barber2003algorithm} and decomposing the lower bound into a per-timestep reward that helps assign credit for learning the exploration policy $\pi$.
Below, we derive this for the case of learning only a single exploration policy $\pi$ to uncover all errors, though we will later discuss how we can factorize this to learn $N$ exploration policies $\{\pi_i\}_{i = 1}^N$ that each explore to uncover a single error type.
\begin{alignat}{2}
    I(\tau; y) &= H[y] - H[y \mid \tau] \\
    &= H[y] + \mathbb{E}_{\program \sim p(\program), \tau \sim \pi(\program)} \left[ \log p(y \mid \tau) \right] \\
    &\geq H[y] + \mathbb{E}_{\program \sim p(\program), \tau \sim \pi(\program)} \left[ \log g(y \mid \tau) \right] \label{eqn:variational_lower_bound} \\
    &= H[y] + \mathbb{E}_{\program \sim p(\program), \tau \sim \pi(\program)} \left[ \log g(y \mid s_0) + \sum_{t = 0}^{T - 1} r_t^\text{exp} \right], \label{eqn:dream} \\
    \text{where } r_t^\text{exp} &= \log g(y \mid \tau_{:t + 1}) - \log g(y \mid \tau_{:t}) \text{ and} \nonumber \\
    \text{$T$ is} &\text{ the length of the trajectory $\tau = (s_0, a_0, r_0, \ldots, s_T)$}. \nonumber \end{alignat}

The inequality in (\ref{eqn:variational_lower_bound}) holds for replacing the true posterior $p(y \mid \tau)$ with any distribution, and (\ref{eqn:dream}) comes from expanding a telescoping series as done by \dream~\citep{liu2021dream}, where $\tau_{:t} = (s_0, a_0, r_0, \ldots, s_t)$ denotes the the trajectory up to the $t^\text{th}$ state.

This derivation provides a shaped reward function $r_t^\text{exp}$ for learning the exploration policy.
Intuitively, this reward captures how much new information the transition $(s_t, a_t, r_t, s_{t + 1})$ 
provides to the feedback classifier $g$ on what errors are in the program:
The reward is high if observing this transition either indicates an error (e.g., the ball enters the goal, but does not score a point) or rules out an error (e.g., the ball enters the goal and scores a point), and is low otherwise.

Additionally, we now have a recipe for maximizing the mutual information $I(\tau ; y)$ to learn both the policy $\pi$ and feedback classifier $g$.
Only the second term in (\ref{eqn:dream}) depends on $\pi$ and $g$.
Hence, we can maximize this lower bound on $I(\tau; y)$ by maximizing the log-likelihood of the label with respect to the feedback classifier $\mathcal{J}_\text{feedback}(g) = \mathbb{E}_{\program \sim p(\program), \tau \sim \pi(\program)} \left[ \log g(y \mid \tau)\right]$,
and maximizing the rewards $r_t^\text{exp}~=~\log g(y \mid \tau_{:t + 1}) - \log g(y \mid \tau_{:t})$ with respect to the policy $\pi$ via reinforcement learning.

\textbf{Factorizing our objective.}
So far, our approach learns a single exploration policy that must uncover all error types in the rubric.
However, learning such a policy can be challenging, especially if uncovering different error types requires visiting very different states.
We instead propose to learn a separate exploration policy $\pi_k$ for each error index of the rubric $k = 1, \ldots, K$.
We can accomplish this by observing that maximizing the mutual information $I((\tau_1, \ldots, \tau_K); y)$ between $K$ trajectories $\tau_i \sim \pi_i$ is also sufficient to maximize the expected grading accuracy.
Furthermore, maximizing the mutual information with each dimension of the label $I(\tau_k; y_k)$, where $\tau_k \sim \pi_k$ for each $k$,
is sufficient to maximize the mutual information with the entire label $I\left((\tau_1, \ldots, \tau_K); y\right)$.
We therefore can derive exploration rewards for each term $I(\tau_k; y_k)$ to learn each policy $\pi_k$ with rewards $r_t^\text{exp} = \log g(y_k \mid \tau_{:t + 1}) - \log g(y_k \mid \tau_{:t})$.
We find that this improves grading accuracy in our experiments (Section~\ref{sec:experiments}) and enables parallel training and testing for the $K$ exploration policies.

\subsection{A Practical Implementation}\label{sec:practical_impl}
\begin{wrapfigure}[8]{r}{.47\textwidth}
    \vspace{-8mm}
    \begin{minipage}{\linewidth}
    \begin{algorithm}[H]
        \footnotesize
        \begin{flushleft}
        \begin{algorithmic}[1]
            \State Sample a training program $\mu$ with label $y$
            \State Roll out policy to obtain trajectory $\tau \sim \pi_k(\program)$
            \State Compute rewards with feedback classifier $r_t^\text{exp} = \log g(y_k \mid \tau_{:t+1}) - \log g(y_k \mid \tau_{:t})$
            \State Update policy to maximize rewards $r_t^\text{exp}$ with RL
            \State Update feedback classifier to max. $\log g(y_k \mid \tau)$
\end{algorithmic}
        \end{flushleft}
        \caption{Training episode for policy $\pi_k$}
        \label{alg:training}
        \vspace{-1mm}
    \end{algorithm}
    \end{minipage}
\end{wrapfigure}

\looseness=-1
Overall, \ours consists of a feedback classifier $g$ and $K$ exploration policies $\{\pi_k\}_{k = 1}^K$, where the $k^\text{th}$ policy $\pi_k$ tries to visit states indicative of whether the $k^\text{th}$ error type is present in the program.
We learn these components by repeatedly running training episodes for each policy $\pi_k$ with Algorithm~\ref{alg:training}.
We first sample a labeled training program and follow the policy on the program (lines 1--2).
Then, we maximize our mutual information objective by updating the policy with our exploration rewards (lines 3--4), and by updating the classifier to maximize the log-likelihood of the label (line 5).

\looseness=-1
In practice, we parametrize the exploration policies and feedback classifier as neural networks.
Since the exploration rewards $r_t^\text{exp}$ depend on the past and are non-Markov, we make each exploration policy $\pi_k$ recurrent:
At timestep $t$, the policy $\pi_k(a_t \mid (s_0, a_0, r_0, \ldots, s_t))$ conditions on all past states, actions, and observed rewards for each $k$.
We parametrize each policy as a deep dueling double Q-networks~\citep{mnih2015human, wang2016dueling, van2016deep}.
Consequently, our policy updates in line 4 consist of placing the trajectory in a replay buffer with rewards $r_t^\text{exp}$ and sampling from the replay buffer to perform Q-learning updates.
We parametrize each dimension of the feedback classifier $g(y_k \mid \tau)$ for $k = 1, \ldots, K$ as a separate neural network.
We choose not to share parameters between the exploration policies and between the dimensions of the feedback classifier for simplicity.
However, we note that significant parameter sharing is likely possible, as exploration policies for two different error types can be extremely similar (e.g., two errors that involve hitting the ball into the goal).
Automatically determining which parameters to share could be an interesting direction for future work, as it is not known a priori which error types are related to each other.
See Appendix~\ref{sec:dream_grader_details} for full architecture and model details.

\subsection{Play-to-Grade as Meta-Exploration}\label{sec:metarl}

Our choice to optimize the mutual information objective $I(\tau; y)$ and decompose this objective into per-timestep rewards using techniques from the \dream meta-RL algorithm stems from the fact that the Play-to-Grade paradigm can be cast as a meta-exploration problem.
Specifically, meta-RL aims to learn agents that can quickly learn new tasks by leveraging prior experience on related tasks.
The standard few-shot meta-RL setting formalizes this by allowing the agent to train on several MDPs (tasks).
At time time, the agent is presented with a new MDP and is allowed to first explore the new MDP for several episodes (i.e., the few shots) to gather information.
Then, it must use the information it gathered to solve the MDP and maximize returns on new episodes.
Learning to efficiently spend these allowed few exploration episodes to best solve the test MDP is the meta-exploration problem.

Our setting of providing feedback follows the exact same structure as few-shot meta-RL.
In our setting, we can view each student program as a new 1-step task of predicting the feedback label, where the reward is the number of dimensions of the label that are correctly predicted.
To predict this label, the Play-to-Grade paradigm first explores the program for several episodes to discover states indicative of errors, which corresponds to the few shots.
The key challenge in this setting is exactly how to best spend those few exploration episodes: i.e., to gather the information needed to predict the label, which is exactly the meta-exploration problem.
This bridge between identifying errors in programs and meta-exploration suggests that techniques from each body of literature could mutually benefit each other.
Indeed, \ours leverages ideas from \dream and future work could explore other techniques to transfer across the two areas.
Additionally, this connection also offers a new testbed for meta-exploration and meta-RL research.
As discussed in Section~\ref{sec:related_work}, while existing meta-RL benchmarks
tend to be either readily accessible or impactful and realistic, automatically providing feedback simultaneously provides both, and we release code for a meta-RL wrapper of the Bounce programming assignment to spur further research in this direction.

\section{Experiments}\label{sec:experiments}

In our experiments, we aim to answer five main questions:
(1) How does automated feedback grading accuracy compare to human grading accuracy?
(2) How does \ours compare with \citet{nie2021play}, the state-of-the art Play-to-Grade approach?
(3) What are the effects of our proposed factorization and derived exploration rewards on \ours?
(4) How much human labor is saved by automating feedback?
(5) Interactive programs can be particularly challenging to grade because test programs can contain behaviors not seen during training --- how well do automated feedback systems generalize to such unseen behaviors?
To answer these questions we consider the dataset of 700K real anonymized Bounce student programs, described in Section~\ref{sec:problem_setting}.

\looseness=-1
Below, we first establish the points of comparison to necessary answer these questions (Section~\ref{sec:baselines}).
Then, we evaluate these approaches to answer the first four questions (Section~\ref{sec:results}).
Finally, we answer question (5) by evaluating \ours on variants of Bounce student programs that modify the ball and paddle speeds, including speeds not seen during training (Section~\ref{sec:creativity}).
Additionally, in Appendix~\ref{sec:additional_results}, we test if \ours can scale to all error types and an additional interactive assignment called Breakout, which is widely taught in university and highschool classrooms.

\subsection{Points of Comparison}\label{sec:baselines}

\looseness=-1
We compare with the following four approaches.
Unless otherwise noted, we train 3 seeds of each automated approach for 5M steps on $N = \num{3556}$ training programs, consisting of $0.5\%$ of the dataset.

\textbf{Human grading.}
To measure the grading accuracy of humans, we asked for volunteers to grade Bounce programming assignments.
We obtained 9 volunteers consisting of computer science undergraduate and PhD students, 7 of whom had previously instructed or been a teaching assistant for a computer science course.
Each volunteer received training on the Bounce programming assignment and then was asked to grade 6 randomly sampled Bounce programs.
See Appendix~\ref{sec:human_details} for details.

\textbf{\citet{nie2021play} extended to provide fine-grained feedback.}
We extend the original Play-to-Grade approach, which provides binary feedback about whether a program is completely correct or not, to provide fine-grained feedback.
Specifically, \citet{nie2021play} choose a small set of 10 training programs, curated so that each program exhibits a single error, and together, they span all errors.
Then, for each training program, the approach learns (i) a distance function $d(s, a)$ that takes a state-action tuple $(s, a)$, trained to be large for tuples from the buggy training program and small for tuples from a correct reference implementation; and (ii) an exploration policy that learns to visit state-actions where $d(s, a)$ is large.
To provide feedback to a new student program, the approach runs each exploration policy on the program and outputs that the program has an error if any of the distance functions is high for any of the tuples visited by the exploration policies.

We extend this approach to provide fine-grained feedback by following a similar set up.
We follow the original procedure to train a separate policy and distance function on $K = 8$ curated training programs, where the $k^\text{th}$ program exhibits only the $k^\text{th}$ error from the rubric we consider.
Then, to provide fine-grained feedback on a new program, we run each policy on the new program and predict that the $k^\text{th}$ error is present (i.e., $\hat{y}_k = 1$) if the $k^\text{th}$ distance function is high on any state-action tuple.
We use code released by the authors without significant fine-tuning or modification.
We emphasize that this approach only uses 8 curated training programs, as opposed to the $N = \num{3556}$ randomly sampled programs used by other automated approaches, as this approach is not designed to use more training programs, and furthermore cannot feasibly scale to many more training programs, as it learns a distance function and policy for every training program.

\textbf{\ours (direct max).}
To study the effect of our derived exploration rewards $r_t^\text{exp}$, we consider the approach of directly maximizing the \ours objective in Equation~\ref{eqn:reparam_grading_accuracy}, described at the beginning of Section~\ref{sec:dream_grader}.
This approach treats the inside of the expectation as end-of-episode returns, and does not provide explicit credit assignment.
This approach is equivalent to maximizing the \ours objective with the \rl meta-RL algorithm~\citep{duan2016rl, wang2016learning}.

\textbf{\ours (unfactorized).}
Finally, to study the effect of our proposed factorization scheme, described at the end of Section~\ref{sec:dream_grader}, we consider a variant of \ours where we do not factorize the objective and instead only learn a single exploration policy to uncover all errors.

\subsection{Main Results}\label{sec:results}

\begin{wraptable}[8]{r}{0.7\textwidth}
    \center\small
    \vspace{-6mm}
    \caption{
        Accuracy, precision, recall and F1 of grading systems, averaged across the $K = 8$ errors of the rubric, with 1-standard deviation error bars.
    }
    \vspace{-1mm}
    \resizebox{0.7\textwidth}{!}{\begin{tabular}{lcccc}
        \toprule
        & Accuracy & Precision & Recall & F1 \\
        \cmidrule(lr){2-2} \cmidrule(lr){3-3} \cmidrule(lr){4-4} \cmidrule(lr){5-5}
        Human & \textbf{95.8 $\pm$ 3.9\%} & \textbf{95.0 $\pm$ 13.2\%} & \textbf{91.1 $\pm$ 10.0\%} & \textbf{91.9 $\pm$ 8.3\%}\\
        \ours & 94.3 $\pm$ 1.3\% & 76.7 $\pm$ 5.8\% & \textbf{94.3 $\pm$ 1.6\%} & 84.6 $\pm$ 1.5\% \\
        \ours (unfactorized) & 91.3 $\pm$ 0.4\% & 72.9 $\pm$ 0.5\% & 68.9 $\pm$ 1.0\% & 70.8 $\pm$ 0.7\% \\
        \ours (direct max) & 84.8 $\pm$ 2.2\% & 36.3 $\pm$ 1.7\% & 37.8 $\pm$ 9.7\% & 36.6 $\pm$ 5.1\% \\
        \citet{nie2021play} & 75.5 $\pm$ 0.9\% & 24.9 $\pm$ 5.0\% & 27.7 $\pm$ 7.1\% & 26.1 $\pm$ 5.7\% \\
        \bottomrule
    \end{tabular}
    }
\label{tab:main_results}
    \vspace{-3.5mm}
\end{wraptable}

We compare the approaches based on grading accuracy, precision, recall, and F1 scores averaged across the 8 error types in the rubric.
Table~\ref{tab:main_results} summarizes the results.
Overall, we find that \ours achieves the highest grading accuracy of the automated grading approaches, providing feedback with 17.7\% greater accuracy than \citet{nie2021play}.
Furthermore, \ours comes within 1.5\% of human-level grading accuracy.
\ours achieves this by learning exploration behaviors that probe each possible error event (see Appendix~\ref{sec:visualizations} or \url{https://ezliu.github.io/dreamgrader} for visualizations of the learned behaviors).

\begin{figure}[t]
    \centering
    \includegraphics[width=\linewidth]{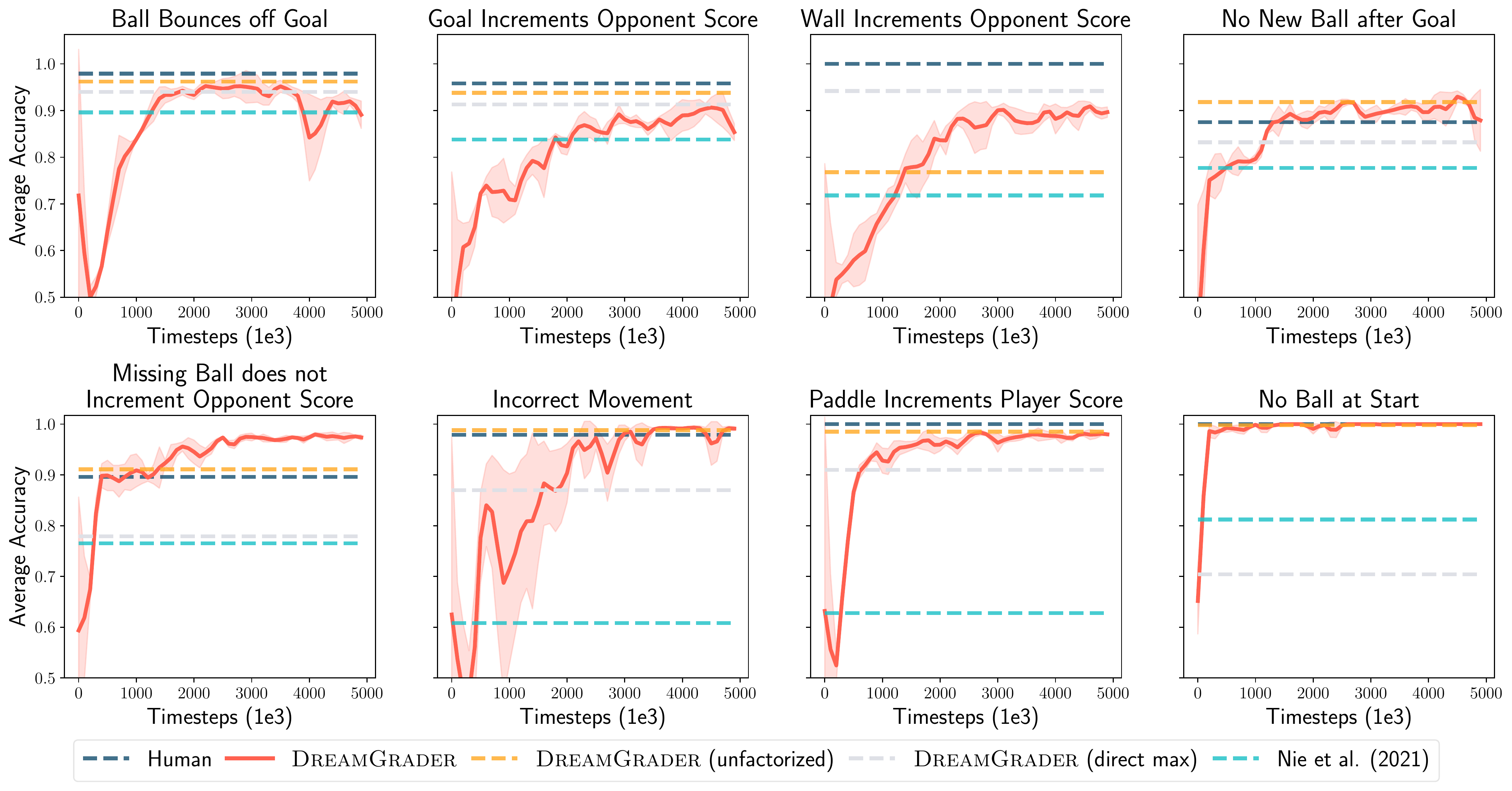}
    \vspace{-3mm}
\caption{Average grading accuracy for each error type vs. number of training steps for \ours with 1-stddev error bars. We plot the final grading accuracies of the other approaches as horizontal lines.}
    \label{fig:learning_curves}
    \vspace{-2mm}
\end{figure}

\looseness=-1
\textbf{Analysis.}
To further understand these results, we plot the training curves of the grading accuracy on each error type in the rubric vs. the number of training steps of \ours, as well as the final grading accuracies of the other approaches in Figure~\ref{fig:learning_curves}.
The performance of variants of \ours underscores the importance of our design choices:
\ours (direct max) achieves significantly lower accuracy than \ours across all error types, indicating the importance of our shaped exploration rewards $r_t^\text{exp}$ for learning effective exploration.
Additionally, while \ours (unfactorized) achieves relatively high average accuracy, it still performs worse than \ours. This illustrates the difficulty of learning a single exploration policy to uncover all errors, which is alleviated by our factorization.

Additionally, we find that \ours achieves human-level grading accuracy on 5 of the 8 error types, but relatively struggles on the 3 errors on the left of the top row of Figure~\ref{fig:learning_curves}, which leads to overall lower average grading accuracy.
For the errors involving incrementing the opponent score, we qualitatively find that \ours most commonly struggles when there are multiple balls, which makes it difficult to ascertain which events are causing the opponent score to increment.
Human grading was able to circumvent this issue, but humans required playing the program for up to 40 episodes, while we limited \ours to a single episode per error type.

\textbf{Reduction in human grading burden.}
We found that our grading volunteers required between 1--6 minutes to grade each program, averaging around 3 minutes.
At this grading speed, grading all of the 700K Code.org submissions would take close to 4 years of human labor.
In contrast, automatic grading with \ours requires only 1 second per program on a single NVIDIA RTX 2080 GPU,
which is $180\times$ faster.

\subsection{Generalizing to Unseen Behaviors}\label{sec:creativity}

\looseness=-1
A key challenge in providing feedback for interactive programs is that student programs may include behaviors not seen during training.
To evaluate generalization to unseen behaviors, we consider varying the speeds of the ball and paddle.
There are five possible settings for the ball and paddle speeds: very slow, slow, normal, fast, and very fast.
To test the ability of \ours to generalize to unseen behaviors at test time, we train \ours on programs where we hold out the normal ball and paddle speeds, and then test \ours on programs with the held out speeds.
Specifically, we use the same $N = \num{3556}$ training programs as before, but we uniformly randomize the ball and paddle speeds independently to be one of the four non-held out speeds.
Then we evaluate on test programs with (1) held out ball and paddle speeds; (2) held out ball speed with random training paddle speed; (3) held out paddle speed with random training ball speed; and (4) training ball and paddle speeds.

\begin{wraptable}[7]{r}{0.7\textwidth}
    \center\small
    \vspace{-6mm}
    \caption{
        \ours's results under held out ball and paddle speeds.
        \ours generalizes to ball and paddle speeds not seen during training.
    }
    \vspace{-2mm}
    \resizebox{0.7\textwidth}{!}{\begin{tabular}{lcccc}
        \toprule
        & Both held out & Held out ball speed & Held out paddle speed & No held out speed \\
        \cmidrule(lr){2-2} \cmidrule(lr){3-3} \cmidrule(lr){4-4} \cmidrule(lr){5-5}
        Accuracy & 89.0 $\pm$ 1.9\% & 89.7 $\pm$ 1.7\% & 89.6 $\pm$ 2.0\% & 89.3 $\pm$ 1.8\% \\
        Precision & 40.0 $\pm$ 1.0\% & 44.1 $\pm$ 1.8\% & 46.2 $\pm$ 1.3\% & 41.6 $\pm$ 2.1\% \\
        Recall & 88.2 $\pm$ 2.3\% & 89.3 $\pm$ 1.5\% & 89.0 $\pm$ 1.7\% & 85.5 $\pm$ 2.4\%\\
        F1 & 55.0 $\pm$ 1.3\% & 59.1 $\pm$ 1.9\% & 60.1 $\pm$ 0.8\% & 56.0 $\pm$ 2.4\% \\
        \bottomrule
    \end{tabular}
    }
\label{tab:invariances}
    \vspace{-3.5mm}
\end{wraptable}

\ours generalizes to unseen ball and paddle speeds at test time.
Compared to the standard training in the previous section, where all ball and paddles had the ``normal'' speed, performance drops, as under faster balls, certain behaviors like hitting the ball into the goal cannot be reliably achieved, and increasing the training data does not decrease the performance drop.
However, accuracy remains relatively high, and \ours performs about the same on test programs regardless of whether the ball and paddle speeds were seen during training or not.
This indicates some ability to generalize to unseen behaviors at test time.
Table~\ref{tab:invariances} displays the full results.

\section{Conclusion}\label{sec:conclusion}
\looseness=-1
In this work, we introduced \ours, an automatic grading system for interactive programs that provides fine-grained feedback at near human-level accuracy.
The key insight behind our system is connecting the problem of automatically discovering errors with the meta-exploration problem in meta-RL, which yields important benefits for both sides.
On the one hand, this connection offers a powerful and previously unexplored toolkit to computer science education, and more generally, to discovering errors in websites or other interactive programs.
On the other hand, this connection also opens impactful and readily accessible applications for meta-RL research, which has formerly primarily focused on synthetic tasks due to the lack of more compelling accessible applications.

\looseness=-1
While \ours nears human-level grading accuracy, we caution against blindly replacing instructor feedback with automated grading systems, such as \ours, which can lead to potentially negative educational and societal impacts without conscientious application.
For example, practitioners must ensure that automated feedback is equitable and not biased against certain classes of solutions that may be correlated with socioeconomic status.
One option to mitigate the risk of potential negative consequences while still reducing instructor labor is to use automated feedback systems to \emph{assist} instructors similar to prior work~\citep{glassman2015overcode}, e.g., by querying the instructor on examples where the system has low certainty, or presenting videos of exploration behavior from the system for the instructor to provide the final feedback.

Finally, this work takes an important step to reduce teaching burden and improve education, but we also acknowledge \ours still has important limitations to overcome.
Beyond the remaining small accuracy gap between \ours and human grading, \ours also requires a substantial amount of training data.
While we only use 0.5\% of the Bounce dataset for training, it still amounts to 3,556 labeled training programs, and labeling this many programs can be prohibitive for smaller-scale classrooms, though feasible for larger online platforms.
We hope that our release of Bounce as a meta-RL problem can help spur future work to overcome these limitations.

\textbf{Reproducibility.} Our code is publicly available at \url{https://github.com/ezliu/dreamgrader}, which includes Bounce as a new meta-RL testbed.

\begin{ack}

We thank Kali Stover from the Institutional Review Board (IRB) and Ruth O'Hara and Tallie Wetzel from the Student Data Oversight Committee (SDOC) for reviewing our process of asking humans to grade programs.
We thank our human graders: Annie Xie, Sahaana Suri, Cesar Lema, Olivia Lee, Moritz Stephan, Kaien Yang, Maximilian Du, Patricia Strutz, and Govind Chada.

We thank Annie Xie and Sahaamble Suri for thwarting EL's best attempts at procrastination, without whom, this work would not have been completed in time.

EL is supported by a National Science Foundation Graduate Research Fellowship under Grant No. DGE-1656518.
CF is a Fellow in the CIFAR Learning in Machines and Brains Program.
This work was also supported in part by Google, Intel, and a Stanford Human-Centered AI Hoffman Yee grant.
Icons in this work were made by FreePik from FlatIcon.

\end{ack}

{
\small

\bibliography{all}
\bibliographystyle{plainnat}
}

\newpage
\appendix

\section{Dataset Details}\label{sec:dataset_details}

We use the Bounce programming assignment dataset from Code.org, released by~\citet{nie2021play}.
We release code that packages this dataset as an easy-to-use meta-RL testbed.

\textbf{Rubric.} Our rubric contains the following 8 error types, spanning all of the events and consequences listed in Table~\ref{tab:error_types}.
\begin{enumerate}
    \item When the ball hits the goal, it incorrectly bounces off.
    \item When the ball hits the goal, the opponent score is incorrectly incremented.
    \item When the ball hits the goal, no new ball is launched.
    \item When the ball hits the floor, the opponent score is not incorrectly incremented.
    \item When the ball hits the wall, the opponent score is incorrectly incremented.
    \item When the left or right action is taken, the paddle moves in the wrong direction.
    \item When the ball hits the paddle, the player score is incorrectly incremented.
    \item When the program starts, no ball is launched.
\end{enumerate}

Some errors in the program make it impossible to uncover other errors.
For example, if no ball is launched at the start of the program, it is impossible to check whether error 7. is present in the program, because there is no ball to hit on the paddle.
In these situations, we label all of the impossible to check errors as not present in the program.

Future work could investigate iteratively providing feedback on a program to a student in order to provide feedback about all errors, including those that are initially impossible to uncover.
Such a system could work as follows:
First, the grading system provides feedback about all of the errors it can currently find in the program.
Then, the student updates their program to fix all of the errors found by the program.
Finally, the student re-submits their updated program to the grading system for another round of evaluation.
This process continues until there are no more errors in the program.

\textbf{Statistics.}
The dataset consists of 711,274 submissions from 453,211 students --- some students created multiple submissions.
Amongst the 711,274 submissions, there are 111,773 unique programs.
The error labels $y$ assigned to each program were programmatically generated by~\citet{nie2021play}.

Averaged across the 6 programs and 9 human graders, humans achieve a raw inter-grader agreement of 85.4\%.
Agreement is computed as the fraction of label dimensions that all 9 human graders agree on.

\textbf{Recommended settings.} Though our code can flexibly support many settings, we recommend the following standard configurations to enable better comparison on future work on this dataset.
We recommend that each episode last for $100$ timesteps and terminate if the player or opponent score exceeds $30$.
We also recommend allowing $K$ episodes for exploring each program for bugs, where $K$ is the number of considered bugs (i.e., the dimensionality of the label).

\section{\ours Details}\label{sec:dream_grader_details}
Our implementation of \ours builds off the \dream code released by~\citet{liu2021dream} at \url{https://github.com/ezliu/dream}.
We use the PyTorch~\citep{paszke2019pytorch} DQN implementation released by~\citet{liu2020learning} and experimental infrastructure from~\citet{liu2020imitation}.

\subsection{Model Architecture}
\ours consists of $K$ recurrent exploration policies $\{\pi_k\}_{k = 1}^K$ and a feedback classifier $g(y \mid \tau)$.

\textbf{Exploration policies.} Each exploration policy $\pi_k$ is parametrized as a double dueling deep Q-network~\citep{mnih2015human,wang2016dueling,van2016deep}, consisting of a recurrent Q-function $Q(\tau_{:t}, a_t)$ and a target network $Q_\text{target}(\tau_{:t}, a_t)$ with the same architecture as the Q-function, where $\tau_{:t} = (s_0, a_0, r_0, \ldots, s_t)$ denotes the trajectory so far up to timestep $t$.
To parametrize these Q-functions, we first embed the trajectory $\tau_{:t}$ as $e(\tau_{:t}) \in \mathbb{R}^{64}$.
Then, we apply two linear layers with output size $1$ and $|\mathcal{A}|$ respectively.
These represent the state-value function $V(\tau_{:t})$ and advantage $A(\tau_{:t}, a_t)$ respectively.
Finally, following the dueling architecture~\citep{wang2016dueling}, we compute the Q-value as:
\begin{equation}
    Q(\tau_{:t}, a_t) = V(\tau_{:t}) + A(\tau_{:t}, a_t) - \frac{1}{|\mathcal{A}|}\sum_{a \in \mathcal{A}}A(\tau_{:t}, a).
\end{equation}

To compute the trajectory embedding $e(\tau_{:t})$, we embed each tuple $(s_{t'}, a_{t'}, r_{t'}, s_{t' + 1})$ for $t' = 0, \ldots, t - 1$ and then pass an LSTM~\citep{hochreiter1997lstm} over the embeddings of the tuples.
The embedding of $(s_{t'}, a_{t'}, r_{t'}, s_{t' + 1})$ is computed by embedding each component and applying a final linear layer with output dimension 64 to the concatenation of the embedded components.
We embed $s_{t'}$ and $s_{t' + 1}$ with the same network, using the architecture from~\citet{nie2021play}, consisting of two linear layers with output dimensions $128$ and $64$, respectively, with an intermediate ReLU activation.
We embed the action $a_t$ with an embedding matrix with output dimension $16$.
We embed the scalar reward $r_t$ with a single linear layer of output dimension $32$.

Each exploration policy $\pi_k$ is trained to maximize the expected discounted exploration rewards via standard DQN updates:
\begin{align}
    \mathcal{J}_\text{exp}(\pi) &= \mathbb{E}_{\program \sim p(\program), \tau \sim \pi(\program)}\left[\sum_{t = 0}^T r_t^\text{exp} \right], \\
    \text{where } r_t^\text{exp} &= \log g(y_k \mid \tau_{:t + 1} - \log g(y_k \mid \tau_{:t}).
\end{align}

\textbf{Feedback classifier.}
The feedback classifier $g(y \mid \tau)$ outputs a distribution over predicted labels $y \in \{0, 1\}^K$.
We parametrize each dimension of the feedback classifier $g(y_k \mid \tau)$ with a separate neural network for simplicity.
To parametrize $g(y_k \mid \tau)$, we embed the trajectory $\tau$ as $e(\tau)$ using a network similar to the one for the exploration policy.
Then, we apply three linear layers with output dimensions 128, 128, and 2 respectively to $e(\tau)$ with intermediate ReLU activations.
Finally, we apply a softmax layer to the output of the linear layers, which forms the distribution over $y_k \in \{0, 1\}$.

To embed the trajectory $\tau$ as $e(\tau)$, we embed each $(s_t, a_t, r_t, s_{t + 1})$-tuple for each timestep $t$ in the trajectory $\tau$.
Then, we pass an LSTM with output dimension 128 over the embeddings of the tuples, and take the last hidden state of the LSTM as $e(\tau)$.
To embed each $(s_t, a_t, r_t, s_{t + 1})$-tuple, we embed each component separately, and then apply two final linear layers with output dimensions 128 and 64 respectively and an intermediate ReLU activation.
We use the same networks architectures used in the exploration policy trajectory embeddings to embed the state, action, and reward components.

Each dimension of the feedback classifier $g(y_k \mid \tau)$ is trained to maximize:
\begin{align}
    \mathcal{J}_\text{feedback}(g) = \mathbb{E}_{\program \sim p(\program), \tau \sim \pi(\program)} \left[ \log g(y_k \mid \tau)\right].
\end{align}

\subsection{Hyperparameters}
\begin{table}[]
    \centering
    \caption{Hyperparameters used for \ours.}
    \begin{tabular}{ll}
        \toprule
        Hyperparameter & Value \\
        \midrule
        Discount Factor $\gamma$ & 0.99 \\
        Learning Rate & 0.0001 \\
        Replay buffer batch size & 32 \\
        Replay buffer minimum size before updating & 500 episodes \\
        Target parameters syncing frequency & 5000 updates \\
        Update frequency & 4 steps \\
        Grad norm clipping & 10 \\
        \bottomrule
    \end{tabular}
    \label{tab:hyperparameters}
\end{table}

We use the hyperparameters listed in Table~\ref{tab:hyperparameters} for all of our experiments.
We chose values based on those used in~\citet{liu2021dream} used for \dream, and did not tune these values.
We optimize the objectives written above using the Adam optimizer~\citep{kingma2015adam}.
During training, we anneal the $\epsilon$ for $\epsilon$-greedy exploration from $1$ to $0.01$ over 250,000 steps.
We use $\epsilon = 0$ during evaluations.

\textbf{Delayed updating.}
At the beginning of training, the feedback classifier is randomly initialized and provides poor noisy reward signal for learning exploration.
However, as the exploration policy occasionally visits states indicative of bugs, the feedback classifier begins to learn and improve its reward signal for the exploration policy.
To ensure that the exploration policy only learns from good reward signal, we wait until the replay buffer contains at least 500 episodes before beginning to take gradient steps on the exploration policy, which is common in DQN-based algorithms.
This allows time for the feedback classifier to learn before the exploration policy begins updating on the reward computed from the feedback classifier.

\section{Experiment Details}\label{sec:experiment_details}
\subsection{Human Grading}\label{sec:human_details}

\textbf{Grading details.}
We obtained 9 volunteers to measure the grading accuracy of humans by soliciting volunteers from a university research group, consisting of computer science undergraduate, master's and PhD students involved in machine learning research.
Each volunteer first received training about the Bounce programming assignment by reading a document describing the behavior of a correct Bounce program and all potential errors that may occur in incorrect implementations.
Additionally, each volunteer was allowed to play a correct implementation for as long as they desired.
After receiving training, each volunteer was presented with the same set of 6 randomly sampled Bounce programs, but in a randomized order.
For each program, each volunteer was allowed to play the program for as long as they needed and was asked to list the errors they found in the program on a checklist including all possible errors.
The reported human grading accuracy in Table~\ref{tab:main_results} is the mean accuracy of the 9 volunteers on these 6 programs.

\textbf{Compensation.} Volunteers took 30--45 minutes total to familiarize themselves with the Bounce programming assignment and to grade the 6 programs.
They were each compensated with a \$10 gift card, amounting to a compensation of roughly \$15 per hour.

\textbf{Institutional Review Board.} The grading process did not expose grading volunteers to any risks beyond that of normal life.
It was reviewed by the Institutional Review Board (IRB) and it was determined that it did not constitute human subjects research and did not require IRB approval.
Below is the final determination of the IRB.

\begin{displayquote}
    After further review, the IRB has determined that your research does not involve human subjects as defined in 45 CFR 46.102(f) and therefore does not require review by the IRB.
\end{displayquote}

\section{Additional Results}\label{sec:additional_results}
\subsection{Results on Additional Error Types}
Our experiments primarily focus on 8 error types that span all event and consequence types in Table~\ref{tab:error_types} for simplicity.
However, below, we include results of \ours on all error types for completeness.
There are 28 total error types, corresponding to the 6 event types times the 5 consequence types, minus 2 of the pairs --- ``when the paddle moves, the ball bounces / does not bounce'' and ``when the program starts, the ball bounces / does not bounce''  --- which are not found in student programs.

We train \ours on the same $N = \num{3556}$ programs as the other experiments, but train and evaluate on all 28 error types. 
We find that \ours's performance remains roughly the same as when it is trained and evaluated on only 8 error types.
Specifically, \ours achieves an average accuracy of 94\% on all 28 error types, whereas it achieves an average accuracy of 94.3\% on only 8 error types, and human accuracy is 95.8\%.
We did not find these results to be surprising, as the original 8 error types were already relatively representative, so we expected performance of the other error types to be similar.

\newpage

\subsection{Results on Breakout}\label{app:breakout}

\begin{wrapfigure}[14]{r}{.4\textwidth}
    \vspace{-4mm}
    \centering
    {
    \setlength{\fboxrule}{2pt}
    \fbox{\includegraphics[width=0.3\textwidth]{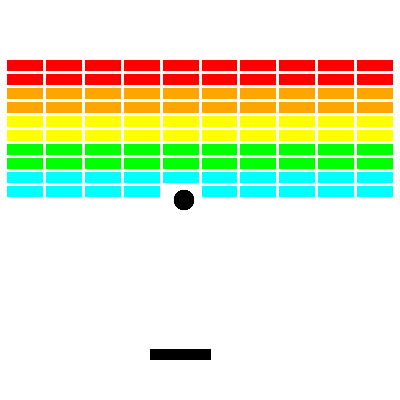}}
    }
    \caption{Breakout. One of the bricks was hit by the ball, creating a gap.}
    \label{fig:breakout}
\end{wrapfigure}

In this section, we include results on the Breakout assignment, an assignment widely taught in university and highschool classrooms.
In this assignment, students are asked to program a Python game, which contains a ball and paddle, as well as several rows of bricks.
The paddle should be able to move left or right, and the ball should bounce off the paddle, walls, and bricks.
If the ball hits a brick, the brick should disappear.
See Figure~\ref{fig:breakout} for an example program.

We did not have access to real student programs.
However, we have the true rubric used to grade this assignment, which we have used to create 64 programs that include common student errors.
We train \ours on 32 of these programs and hold out the remaining 32 for testing.
These errors include:
\begin{itemize}
    \item \emph{Paddle skewer.} The ball always reverses y-directions when it hits the paddle. This makes it possible to "skewer" the ball on the paddle, if the paddle hits the ball from the side as it is falling. When the ball makes contact with the paddle, it reverses directions, but may not move far enough in a single timestep to escape the paddle. Then, on the next timestep, it reverses directions again. This continues and the ball becomes stuck "skewered" on the paddle. In a correct implementation, the ball should only reverse y-directions when it hits the paddle if the ball is falling (and not if it is rising).
\item \emph{Does not delete brick.} The ball bounces off bricks, but the bricks do not disappear when hit.
    \item \emph{Does not bounce off brick.} The ball incorrectly does not bounce off the bricks.
    \item \emph{Wrong number of brick rows.} The assignment specifies that students should create 10 rows of bricks (to test their ability to program loops). It is incorrect if there are any number of rows of bricks not equal to 10.
    \item \emph{Reversed paddle movement.} Pressing the left arrow key results in the paddle moving in any direction other than left, and similarly for the right arrow key.
    \item \emph{Bounce off floor.} The ball incorrectly bounces off the floor, rather than falling out of screen.
\end{itemize}

We test \ours on the error type that is most challenging for humans to grade, \emph{paddle skewer}.
This is challenging to grade because it requires careful timing to hit the ball on the side.
We find that \ours achieves an accuracy of 93.8\% on the test programs, correctly grading 30 of the 32.
\ours learns to deliberately skewer the ball with the paddle, which can be seen at \url{https://ezliu.github.io/dreamgrader}.
We also find that \ours's performance is even higher on the less complex error types, and achieves 96.9\% accuracy on the test programs for the \emph{does not delete brick} error type.

\textbf{Experimental details.} Each Breakout episode consists of 300 timesteps and terminates early if all of the bricks are destroyed, or if the ball hits the floor. The reward function across all programs is 0 for all states and actions. The state consists of the $(x, y)$-coordinates of the ball, paddle, and bricks.
We use the same \ours architecture and hyperparameters that were used for the Bounce experiments.

\begin{figure}[t]
    \centering
    \includegraphics[width=\linewidth]{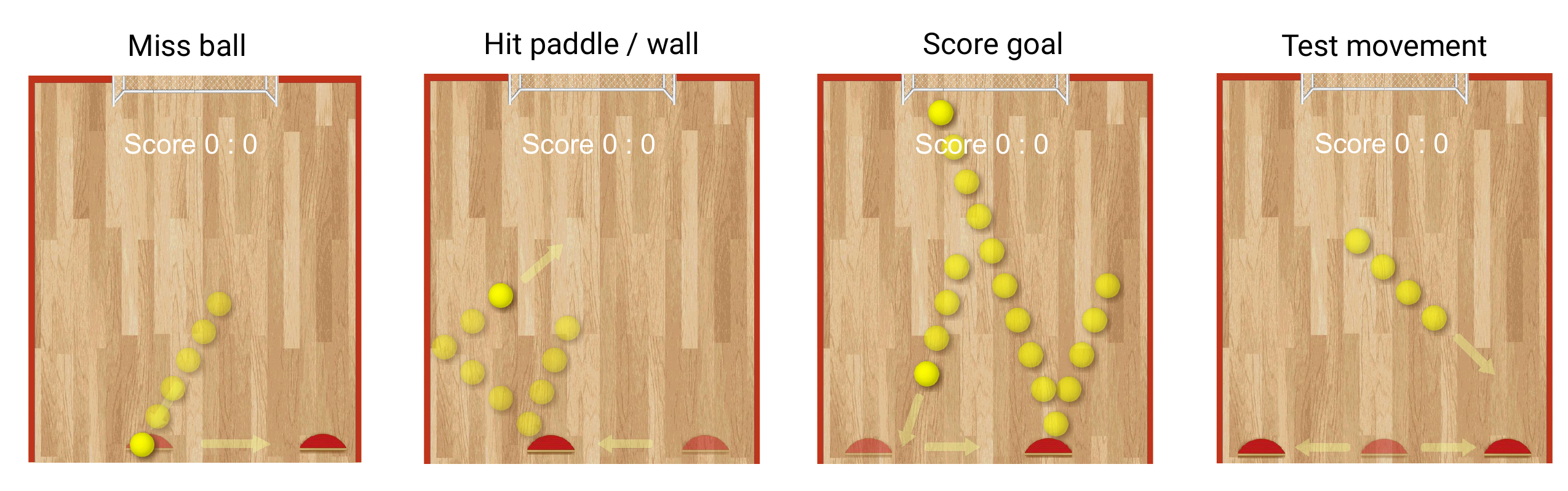}
    \caption{The exploration behaviors learned by \ours. These exploration behaviors probe all of the possible events associated with errors, enabling \ours to uncover all errors.
    Crucially, these exploration behaviors are robust to different programs.
    For example, the illustrated ``score goal'' exploration behavior succeeds in hitting a ball into the goal, even when there are multiple balls.}
    \label{fig:exploration_visualization}
\end{figure}

\subsection{Bounce \ours Exploration Behavior Visualizations}\label{sec:visualizations}
We visualize the exploration behaviors learned by \ours's exploration policies in Figure~\ref{fig:exploration_visualization}.
Qualitatively analyzing these behaviors shows that \ours exploration behaviors that probe each possible event type listed in Table~\ref{tab:error_types}.
Specifically, \ours learns to hit the ball into the goal; hit the ball into the wall; hit the ball with the paddle; deliberately miss the ball; and move the paddle in various directions.
Crucially, we find that behaviors are fairly robust to different programs.
For example, we find that the exploration policies for bugs related to hitting the ball into the goal still successfully hit the ball into the goal most of the time, even when there are multiple balls, or when the actions are reversed, so that the left action moves the paddle right, and vice-versa.
This indicates that \ours can handle the key challenge of our problem setting: learning diverse and adaptable exploration behaviors.
Videos of \ours's exploration behavior can also be found at \url{https://ezliu.github.io/dreamgrader}.

\section{Detailed Relation to Prior Work}\label{sec:detailed_prior_work}
\subsection{Connection with \dream}

In this section, we discuss the full connection between \ours and \dream.
\dream is a generic few-shot meta-RL algorithm that learns two components, an exploration policy, and an exploitation policy.
Assuming that each of the meta-training tasks are distinguishable from each other with a unique problem ID (e.g., a unique one-hot), the \dream exploitation policy learns to solve each task conditioned on an information-bottlenecked representation of the problem ID $z$, which aims to identify only the task-relevant information.
Then, the exploration policy is learned to maximize the mutual information between exploration trajectories from rolling out the exploration policy, and bottlenecked representation $z$, which contains only the task-relevant information.

Primarily, \ours leverages the exploration aspect of \ours.
Specifically, \ours learns an exploration policy to maximize the mutual information between exploration trajectories and the label $y$, by using the same per-timestep dense reward decomposition that \dream uses to maximize its mutual information objective.
However, \ours's feedback classifier can also be seen as a special-case of the \dream exploitation policy:
We can interpret exploitation episodes as single timestep episodes, where the feedback classifier takes a single action, i.e., predicting the label, and the reward is given as the number of label dimensions that are correctly predicted, and hence serves the same purpose as the exploitation policy.
The difference is that whereas \dream attempts to learn a representation $z$ that contains only the information needed from exploration by imposing an information bottleneck, $y$ already contains exactly that information, so no such bottleneck needs to be applied.

\subsection{Differences with Original Play-to-Grade Formulation}
In this section, we discuss the primary high-level differences between \ours and~\citet{nie2021play}.
While both approaches aim to learn policies that discover errors in student programs by interacting with them, these approaches conceptually differ in how they frame this problem.
\citet{nie2021play} formulate the feedback challenge as computing the distance between two MDPs, a student program, and a correct reference program, and finding MDPs where this distance is over a threshold, indicating a buggy program.
This leads to an approach, which involves learning a distance function between MDPs, based on how the dynamics and reward functions differ between the two MDPs, and an exploration policy to explore states where the dynamics and rewards significantly differ according to the distance function.
However, \citet{nie2021play} learn this distance function by learning dynamics and rewards models, which can be challenging, and result in poor reward signal for learning the exploration policy.

In contrast, this work connects the feedback challenge with the meta-exploration and meta-RL problem statements, which opens to door for applying meta-RL techniques.
Specifically, \ours leverages techniques from the \dream meta-RL algorithm, which creates a dense and helpful reward signal for learning an exploration policy.
Empirically, this reward signal results in much more effective exploration policies.

Overall, \ours and \citet{nie2021play} significantly conceptually differ in how they approach the feedback problem:
\citet{nie2021play} cast the problem as computing distances between MDPs, whereas \dream casts the problem as meta-exploration, which results in superior performance.

\end{document}